# LEARNING LIGHT FIELD SYNTHESIS WITH MULTI-PLANE IMAGES: SCENE ENCODING AS A RECURRENT SEGMENTATION TASK

*Tomás Völker, Guillaume Boisson, Bertrand Chupeau*

InterDigital Research & Innovation

## ABSTRACT

In this paper we address the problem of view synthesis from large baseline light fields, by turning a sparse set of input views into a Multi-plane Image (MPI). Because available datasets are scarce, we propose a lightweight network that does not require extensive training. Unlike latest approaches, our model does not learn to estimate RGB layers but only encodes the scene geometry within MPI alpha layers, which comes down to a segmentation task. A Learned Gradient Descent (LGD) framework is used to cascade the same convolutional network in a recurrent fashion in order to refine the volumetric representation obtained. Thanks to its low number of parameters, our model trains successfully on a small light field video dataset and provides visually appealing results. It also exhibits convenient generalization properties regarding both the number of input views, the number of depth planes in the MPI, and the number of refinement iterations.

***Index Terms***— Light field, View synthesis, Multi-plane Image (MPI), Deep Learning, Convolutional Neural Network (CNN)

## 1. INTRODUCTION

View synthesis is one of the most popular problems of Computer Vision. Significant innovations have been steered over the last decade by novel hardware features both at acquisition and display sides, from plenoptic cameras to Virtual Reality headsets. Like many other problems, view synthesis also recently experienced dazzling developments with the rise of deep learning. While the first convolutional networks could only infer a single view at a time [1][2], next ones got the ability to generate a whole light field [3][4][5][6]. Among the solutions that address large baseline contents, *Multi-plane Images* stand out as an efficient solution to encode 3D scenes.

A Multi-plane Image (MPI) derives from the *Layered Depth Image* (LDI) [7] and denotes a stack of fronto-parallel layers located at different depths from a reference camera, each layer consisting in a RGBA image. An MPI is a powerful scene description, that can encode diffuse surfaces as well as non-Lambertian effects such as transparencies and reflections. Given an MPI, any novel view can be rendered by a simple homography and alpha-compositing.

Zhou *et al.* proposed in [4] a stereo magnification method, where an MPI inferred from a stereo pair is used to extrapolate further views. The authors also proposed a dedicated dataset of calibrated camera motion clips. Note that although an MPI has three dimensions, Zhou uses 2D convolutions. Therefore the number of depth planes in the MPI is fixed by the network architecture.

Addressing the same use-case and using the same dataset, Srinivasan *et al.* built upon Zhou's work and proposed a two-step pipeline [5]. Stating that the color layers estimation is strongly ill-posed, they generate an initial MPI that is carved from its occluded content, then re-processed to constrain the final RGB values in a back-to-front order. This prevents occluded areas to be filled with foreground colors, a common artifact in Zhou's results. The authors introduced 3D convolution layers, so that the depth range can vary. They also proposed a novel randomized-resolution training method in which the network is fed with tensors presenting alternately low pixel resolution and high depth resolution and conversely, in order to overcome memory limitations at training time.

Last, inspired by recent advances in Learned Gradient Descent (LGD) [8], Flynn *et al.* formulated the problem of view synthesis as an inverse problem, with the MPI as the model parameters and the input images as the observations. Their DeepView method [6] extends the direct MPI supervision on views to an iterative optimization algorithm which is supposed to learn to regularize the solution from the data itself. Instead of actual gradients, the optimization algorithm is fed with so-called *gradient components*, *i.e.* specific features the actual gradient is function of, and learns to use them to improve the current solution. Practically, these gradient components consist of visual clues that are computed from the current MPI state and the input data. These visual clues are chosen to model the interaction of distanced planes, something a simple feedforward network could not achieve and would require a large receptive field. The DeepView algorithm consists of four iterations, using a 2D convolutional network with the same architecture but different weights at each time. The network also presents a permutation-invariant architecture to process the features of each view, alike point cloud deep learning methods [9]. Training this architecture for perceptual loss against a dataset proposed by the authors yields state-of-the-art results.

In this paper we build upon the DeepView approach with the concern of designing a lightweight network that can be trained on a limited dataset. Our model proves versatile and generalizes successfully beyond the training conditions.

## 2. LEARNING 3D SCENE SEGMENTATION

### 2.1. Multi-plane Images and Plane-sweep Volumes

A Multiplane Image (MPI) is a volumetric representation consisting of layers of fronto-parallel RGBA images located at different depths from a reference camera. Layers are spaced linearly in disparity, *i.e.* in $1/z$.

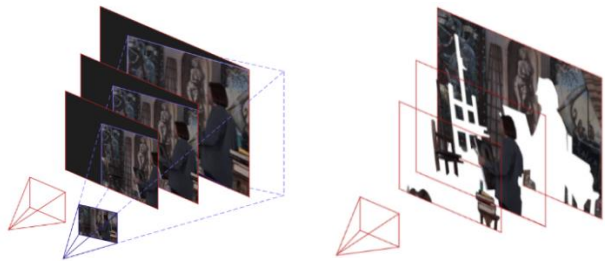

**Figure 1:** Plane-sweep Volume (left) and Multi-plane Image (right). The reference camera is depicted in red.

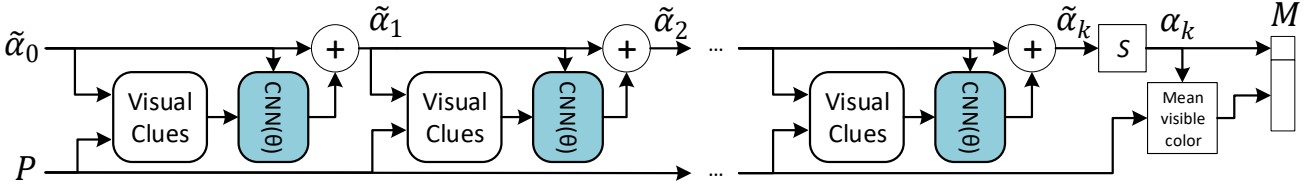

**Figure 2:** Recurrent alpha refiner model. All convolutional networks use the same weights, producing the same function.

An MPI is homogeneous with a *Plane-sweep Volume* (PSV) [10], *i.e.* the collection of images generated when warping an input view onto the reference camera through different depth planes (see Figure 1), which is a common way to ingest calibration data into a convolutional network when looking for correspondences.

### 2.2. Definitions and notations

#### 2.2.1. Camera model and correspondences

For simplicity let's consider contents that have been cleared from distortions. Cameras are therefore modeled as plain pinholes. Let $K_r \in \mathbb{R}^{3\times 3}$ denote the intrinsic matrix of the reference camera – potentially a virtual camera. Let $K$ be the intrinsic matrix of another camera – an actual one –, and $[R \quad T] \in \mathbb{R}^{3\times 4}$ its pose matrix in the coordinate system of the reference camera.

Let's consider a pixel $(u, v)$ in the reference view, and the corresponding voxel located at depth $z$ in the coordinate system of the reference camera. If we denote $z'$ the depth of this voxel in the coordinate system of the other camera, the matching pixel $(u', v')$ in the other view is given by:

$$\begin{bmatrix} u' \\ v' \\ 1 \end{bmatrix} \frac{z'}{z} = H_z . \begin{bmatrix} u \\ v \\ 1 \end{bmatrix} \quad (1)$$

Where the homography matrix $H_z$ is defined as:

$$H_z = K.R^t.K_r^{-1} - \frac{1}{z}[0_{3\times 2} \quad K.R^t.T]$$

#### 2.2.2. From input images to PSVs

We denote *Image Stack* the four-dimensional array corresponding to input views $I = (I_{n,u,v,c}) \in \mathbb{R}^{N\times W\times H\times 3}$. Here for simplicity we assume that every view has the same pixel resolution $W \times H$; $N$ is the number of views, and $c$ denotes the color channel.

We define the broadcasting operator $\mathcal{B}_z : \mathbb{R}^{N\times W\times H\times 3} \to \mathbb{R}^{N\times W\times H\times D\times 4}$ as tiling the image stack along a new depth dimension and adding a mask channel with value 1:

$$\forall d \in \{1, \dots, D\} \quad \mathcal{B}_z(I)_{n,u,v,d,c} = \begin{cases} I_{n,u,v,c} & c \in \{1,2,3\} \\ 1 & c = 4 \end{cases}$$

By convention, depth planes are back-to-front ordered.

We also define a warping operator $\mathcal{W}: \mathbb{R}^{N\times W\times H\times D\times 4} \to \mathbb{R}^{N\times W\times H\times D\times 4}$ which takes the result of broadcasting the image stack and warps the images at different planes using the homography described in equation (1). We assume the array is padded with zeros for the warped coordinates that fall out of the domain. Thus the fourth channel contains a mask equal to 1 if the warped coordinates fall within the image frame and to 0 otherwise.

Eventually we define a *PSV Stack* as the collection of PSVs from multiple views with respect to a certain reference:

$$P = (P_{n,u,v,d,c}) = (\mathcal{W} \circ \mathcal{B}_z)(I)$$

#### 2.2.3. View synthesis: from an MPI back to images

Let's consider an MPI as a four-dimensional array $M = (M_{u,v,d,c}) \in \mathbb{R}^{W\times H\times D\times 4}$. The rendering operation involves a broadcasting step $\mathcal{B}_n$ that operates on views, then the inverse warping $\mathcal{W}^{-1}$ to map each slice onto the desired cameras:

$$\widehat{M} = (\widehat{M}_{n,u,v,d,c}) = (\mathcal{W}^{-1} \circ \mathcal{B}_n)(M)$$

Thus we obtain a five-dimensional image stack, *i.e.* one four-dimensional array per view that is then combined using alpha compositing [11] from back to front.

Omitting the $(n, u, v)$ indices for clarity, we denote an alpha slice $\hat{\alpha}_d = \widehat{M}_{d,c=4}$ and an RGB slice $\hat{I}_d = \widehat{M}_{d,c\leq 3}$. Denoting $\hat{I}_d^A$ and $\hat{\alpha}_d^A$ respectively the RGB and alpha components of the accumulated composed image result, the final rendered image is given by:

$$\hat{\alpha}_D^A . \hat{I}_D^A = \sum_{d=1}^{D} \hat{\alpha}_d . \left( \prod_{i=d+1}^{D} (1 - \hat{\alpha}_i) \right) . \hat{I}_d$$

#### 2.2.4. Voxel visibility

Last, we define the visibility $V$ and the corresponding operator $\mathcal{V}$ as:

$$\begin{cases} V_d = \prod_{i=d+1}^{D} (1 - \hat{\alpha}_i) \\ V = (V_{n,u,v,d}) = \mathcal{V}(\hat{\alpha}) = \mathcal{V}(\widehat{M}_{c=4}) \end{cases}$$

Note that each operator $\mathcal{B}_z$, $\mathcal{B}_n$, $\mathcal{W}$, $\mathcal{W}^{-1}$ and $\mathcal{V}$ is differentiable and therefore enables gradient back-propagation.

### 2.3. Proposed method

Like [6] we consider the problem of generating an MPI as an inverse problem. Input views are considered as the result of the rendering operation from the MPI, and the function to be learned is the inverse of this rendering.

#### 2.3.1. Scene encoding as a volumetric segmentation task

The *Spaces* dataset used in [6] has not been shared, and 4D light fields publicly available for training remain scarce. To overcome this lack of data, we propose a lightweight model whose training requires less iterations and less data.

To this end, we focus on the geometry and therefore estimate only the alpha component of the MPI: $\alpha = (\alpha_{u,v,d}) = M_{c=4}$. This divides by four the amount of values to be predicted, and the scene representation comes down to a segmentation task: encoding the voxels opacity into an alpha volume.

#### 2.3.2. Deriving color layers from the encoded geometry

Once the scene geometry is encoded within the MPI alpha layers, recovering the RGB information is straightforward. Each voxel is assigned to its *mean visible color*, a weighted average of the PSVs by the view visibility:

$$\mu_{u,v,d,c} = \frac{1}{\overline{V}_{u,v,d}} \sum_{n=1}^{N} V^*_{n,u,v,d} P_{n,u,v,d,c}$$

Where $V^*$ denotes the MPI referenced visibility, that indicates the visibility of each voxel with respect to each view: $V^* = (V^*_{n,u,v,d}) = \mathcal{W}(V) = (\mathcal{W} \circ \mathcal{V} \circ \mathcal{W}^{-1} \circ \mathcal{B}_n)(\alpha)$.

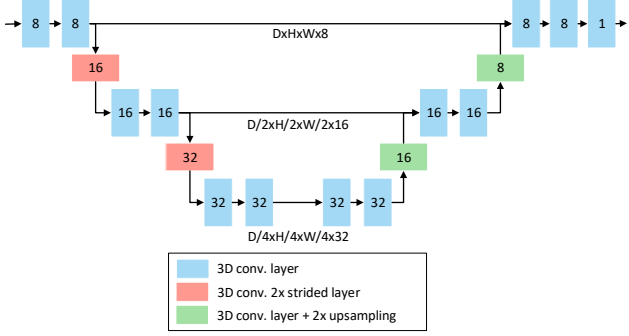

**Figure 3:** U-Net architecture of the refiner network.

And $\overline{V}$ denotes the *total visibility*, a measure of how many actual cameras directly observe the voxel:

$$\overline{V}_{u,v,d} = \sum_{n=1}^{N} V^*_{n,u,v,d}$$

*2.3.3. Geometry refinement with Learning Gradient Descent*

Like in [6] we implement an iterative MPI refinement scheme using the Learning Gradient Descent (LGD) method [8]. Such an optimization framework is expected to learn to regularize the solution directly from the training set. Practically, an LGD framework resembles a residual architecture [12] augmented with visual clues. Given the problem tackled, we consider the following visual clues: the total visibility $\overline{V}_{u,v,d}$, the mean visible color $\mu_{u,v,d,c}$ and the *visible color variance* $\sigma^2_{u,v,d,c}$ defined as:

$$\sigma^2_{u,v,d,c} = \frac{1}{\overline{V}_{u,v,d}} \sum_{n=1}^{N} V^*_{n,u,v,d} \cdot (\mu_{u,v,d,c} - P_{n,u,v,d,c})^2$$

Intuitively, low variance values indicate consensus between cameras, and should therefore induce the creation of new opaque or semi-opaque voxels, until both the variations of the total visibility and the mean visible color become consistent with each other.

Note that every visual clue is reduced along the view dimension. This induces significant memory savings compared to DeepView [6] whose gradient components consist of the PSVs of the input images $P_{n,u,v,d,c}$, the PSVs of the reconstructed images $\hat{P}_{n,u,v,d,c}$, the accumulated over $\hat{\alpha}^A_{n,u,v,d} \cdot \hat{I}^A_{n,u,v,d}$, and the visibilities $V^*_{n,u,v,d}$.

*2.3.4. Recurrent segmentation*

Our scene encoding method is a recurrent scheme, not in the sense of [13] where a convolutional LSTM cell loops over the depth axis, but as the same function is used to refine successively the estimated geometry. Unlike DeepView that uses the same architecture with different weights at each iteration, learning four different successive functions [6], we use the same model with the same weights for each refinement step (see Figure 2). In this sense we train a recurrent geometry refiner model that learns to improve the MPI with respect to its consistency with the PSVs provided, regardless of the final number of refinement iterations.

*2.3.5. Implementation details*

As depicted in Figure 2, the refiner model iteratively updates $\tilde{\alpha}_k$ values that are not restricted in range. Then a sigmoid function is applied to constrain the final $\alpha$ values within $[0; 1]$.

We use a U-Net [14] like network (see Figure 3) similar to the ones used in [5] and [15]. The Table 1 recaps the details of each layer.

| Layer | activ. | kernel | stride | channels | input |
|---|---|---|---|---|---|
| conv1_1 | ReLU | 3×3×3 | 1×1×1 | 8/8 | features |
| conv1_2 | ReLU | 3×3×3 | 1×1×1 | 8/8 | conv1_1 |
| conv1_3 | ReLU | 3×3×3 | 2×2×2 | 8/16 | conv1_2 |
| conv2_1 | ReLU | 3×3×3 | 1×1×1 | 16/16 | conv1_3 |
| conv2_2 | ReLU | 3×3×3 | 1×1×1 | 16/16 | conv2_1 |
| conv2_3 | ReLU | 3×3×3 | 2×2×2 | 16/32 | conv2_2 |
| conv3_1 | ReLU | 3×3×3 | 1×1×1 | 32/32 | conv2_3 |
| conv3_2 | ReLU | 3×3×3 | 1×1×1 | 32/32 | conv3_1 |
| conv3_3 | ReLU | 3×3×3 | 1×1×1 | 32/32 | conv3_2 |
| conv3_4 | ReLU | 3×3×3 | 1×1×1 | 32/32 | conv3_3 |
| conv2_4 | ReLU | 3×3×3 | 1×1×1 | 32/16 | conv3_4 |
| blup2 |  |  |  | 16/16 | conv2_4 |
| conv2_5 | ReLU | 3×3×3 | 1×1×1 | 32/16 | conv2_2 \| blup2 |
| conv2_6 | ReLU | 3×3×3 | 1×1×1 | 16/16 | conv2_5 |
| conv1_4 | ReLU | 3×3×3 | 1×1×1 | 16/8 | conv2_6 |
| blup1 |  |  |  | 8/8 | conv1_4 |
| conv1_5 | ReLU | 3×3×3 | 1×1×1 | 16/8 | conv1_2 \| blup1 |
| conv1_6 | ReLU | 3×3×3 | 1×1×1 | 8/8 | conv1_5 |
| conv1_7 | - | 3×3×3 | 1×1×1 | 8/1 | conv1_6 |

**Table 1:** Convolutional network details. Here "blup" indicates 2× bilinear up-sampling along the three dimensions.

The network contains slightly less than 200K parameters, which is significantly lower than state-of-the-art solutions (millions of parameters in [4], [5] and [6]).

Every layer but the last one is followed by a Rectified Linear Unit (ReLU) activation. The last layer is purely linear to enable unrestricted variations on $\tilde{\alpha}$. The $\alpha_0$ volume is initialized as an empty geometry with an opaque background.

### 2.4. Intrinsic properties

The MPIs generated do not depend on the order of input views. This results from the choice of the visual clues used for LGD. Indeed, both mean visible color, visible color variance and total visibility are view permutation invariant. Also, the model can process any amount of views, potentially more than it had while training, up to the hardware memory limitations.

Last, using 3D convolutions, unlike Zhou [4] and Flynn [6], our model is naturally translational equivariant, *i.e.* translated visual features along the $x$, $y$, or $z$ axis result in a translated geometry.

### 3. EXPERIMENTS

### 3.1. Training

The training is performed with the Technicolor/InterDigital light field dataset [16], augmented with new light field sequences shot with the same 4×4 camera rig. Scenes with objects closer than 1.5m are excluded, resulting in a total of 27 sequences. Due to memory limitations, images are subsampled to 512×272, which corresponds to the resolution addressed in state-of-the-art solutions. We perform online training (batch size of 1), like [4] and [5].

At each training iteration, we first select a random sequence from the dataset, a random frame in the sequence, and a random 3×3 sub-rig. Then we pick the number of views $N$ between 2 and 5. The remaining $(9 - N)$ views are used as target images to supervise the view synthesis. The number of depth planes $D$ of the PSVs and the MPI is chosen randomly between 40 and 50, with $z_1 = +\infty$ and $z_D = 1.5m$. This ensures that the disparity between two successive slices does not exceed one pixel. As for the reference camera, we average the positions and orientations [17] of the input cameras.

During training, the model predicts the MPI using up to 4 refinement steps, starting with 2 and incrementing to 3 and 4 after 20K and 40K iterations respectively, for stability purposes.

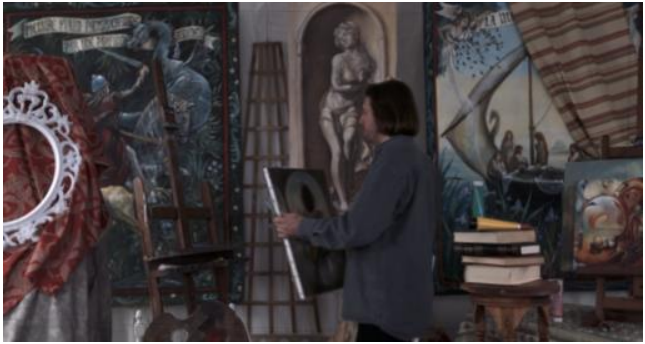
SSIM=0.9382   PSNR=32.38   MAE=0.01589

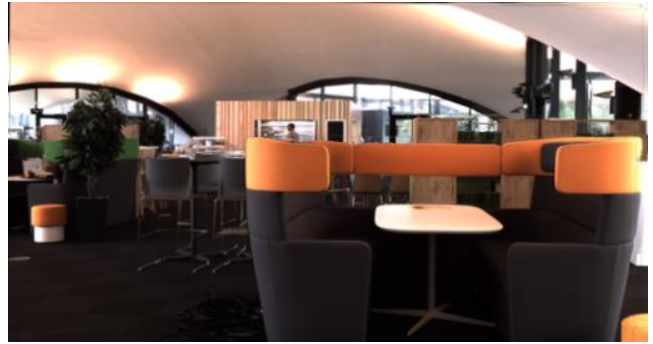
SSIM=0.9435   PSNR=27.26   MAE=0.01735

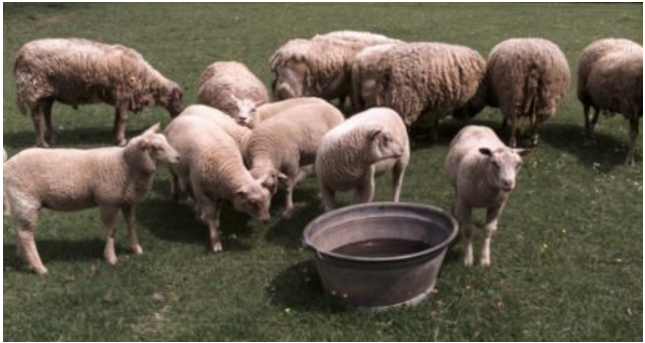
SSIM=0.8904   PSNR=31.22   MAE=0.0195

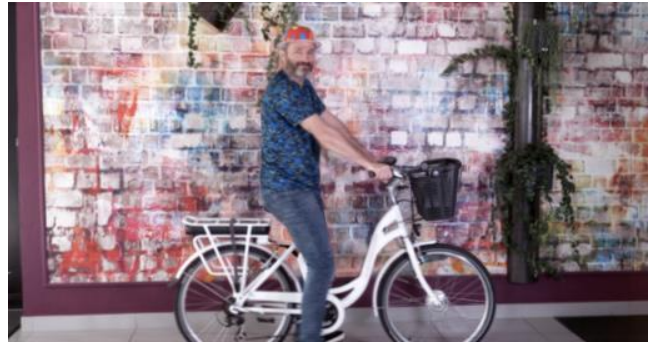
SSIM=0.9119   PSNR=28.62   MAE=0.02469

**Figure 4:** Test inferences. Central views synthesized from an MPI computed from four corner views with four iterations.

The training is supervised on the Structural Similarity (SSIM) measure. We use Instance Normalization [18] and the ADAM optimizer with parameters $\beta_1 = 0.9, \beta_2 = 0.999$ and a learning rate of $10^{-4}$. The model is implemented with TensorFlow. On a NVIDIA Tesla P100 GPU, a training of 200K iterations lasts slightly less than 4 days.

### 3.2. Results

Once trained, the model is evaluated on the 12 sequences of the test dataset. We assess the model by providing the four corner views of a 3×3 sub-rig using 50 depth planes, and we measure the image reconstruction error of the center image using Peak Signal-to-Noise Ratio (PSNR), Structural Similarity (SSIM) and Mean Absolute Error (MAE). The Figure 4 provides visual results examples together with corresponding metrics. Synthesized images are notably artifact-free, though slightly fuzzier than the ground truth.

The Table 2 provides metrics that are averaged along the twelve sequences of the test dataset. As expected, the quality of synthesized images increases at each iteration.

| # iteration | PSNR | SSIM | MAE |
| --- | --- | --- | --- |
| 1 | 28.32 | 0.8670 | 0.02527 |
| 2 | 29.66 | 0.9033 | 0.02104 |
| 3 | 29.83 | 0.9078 | 0.02056 |
| 4 | **29.86** | **0.9084** | **0.02047** |

**Table 2:** Number of refinement steps and image quality.

The Figure 5 shows magnified error maps that correspond to the central view of *Painter* light field. It illustrates the versatility of our model, both in terms of input views (top row) as of refinement steps (bottom row). While the model cannot solve some ambiguities with only two views (see top-left map, *e.g.* the canvas), considering more views (potentially more than the network was trained on) yields a more accurate geometry, therefore better final images.

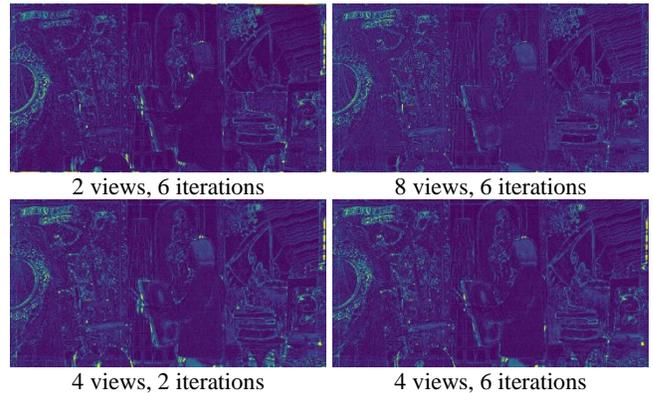

2 views, 6 iterations   8 views, 6 iterations
4 views, 2 iterations   4 views, 6 iterations

**Figure 5:** Magnified error maps on Painter's central view.

## 4. CONCLUSION

A novel light field synthesis scheme based on a recurrent MPI refinement has been proposed. Our main achievement lies in the simplicity and the versatility of our model with respect to the prior art. Contributions to this end are basically twofold. First, we do not address the estimation of color layers but only consider the underlying geometrical segmentation problem. Second, unlike previous approaches we implement a recurrent refinement: the same network loops on successively enhanced MPIs. This yield to a lightweight model (200K parameters *vs.* millions in prior art) that is successfully trained on a reduced dataset consisting of a few dozen light field sequences. The resulting iterative MPI refinement scheme provides visually appealing inferences, while exhibiting nice generalization properties regarding both the number of input views, the number of depth planes in the MPI, and the number of refinement iterations.